\documentclass[10pt,twocolumn,letterpaper]{article}
\usepackage[pagenumbers]{cvpr}
\usepackage{graphicx}
\usepackage{amsmath,amssymb}
\usepackage{booktabs}
\usepackage{colortbl}
\usepackage{multirow}
\usepackage{xcolor}
\usepackage{microtype}
\usepackage{url}
\definecolor{cvprblue}{rgb}{0.21,0.49,0.74}
\definecolor{TableSource}{HTML}{E3EED9}
\definecolor{TableExternal}{HTML}{E4DBF0}
\definecolor{TableHuman}{HTML}{F18B56}
\definecolor{TableCorrect}{HTML}{00A8D9}
\definecolor{TableHeader}{HTML}{F1F1F1}
\usepackage[breaklinks,colorlinks,allcolors=cvprblue]{hyperref}
\newcommand{\nll}{\mathrm{NLL}}
\newcommand{\human}{\mathrm{H}}
\newcommand{\correct}{\mathrm{G}}
\newcommand{\base}{\mathrm{B}}
\newcommand{\decision}{\mathrm{D}}

\usepackage{stfloats}
\usepackage{flushend}
\AddToHook{para/end}{%
  \ifdim\baselineskip=12pt\relax
    \baselineskip=12pt plus .35pt minus .16pt\relax
  \fi
}
\makeatletter
\renewcommand\paragraph{\@startsection{paragraph}{4}{\z@}{1.2ex \@plus .4ex \@minus .2ex}{-1em}{\normalfont\normalsize\bfseries}}
\makeatother

\newcommand{\CommonInternalWrongFav}{36}

\newcommand{\VisualSwapExternalZero}{18}
\newcommand{\VisualMaskExternalNegative}{10}
\newcommand{\VisualMaskExternalPositive}{3}
\newcommand{\VisualMaskExternalZero}{5}

\title{\resizebox{0.98\textwidth}{!}{HumanDec: Are System One Decision Models Better Predictors of Human Visual Choices?}}
\author{Dengzhe Hou\\Tohoku University\\[0.25ex]{\small \href{https://github.com/dengzhe-hou/HumanDec}{\raisebox{-0.43ex}{\includegraphics[page=1,height=2.02ex]{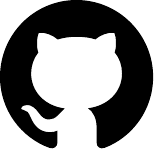}}\hspace{0.31em}GitHub}\enspace\textperiodcentered\enspace\href{https://dengzhe-hou.github.io/HumanDec/}{\raisebox{-0.43ex}{\includegraphics[page=2,height=1.97ex]{figures/platform_icons.pdf}}\hspace{0.31em}Homepage}\enspace\textperiodcentered\enspace\href{https://huggingface.co/datasets/dengzhe-hou/HumanDec}{\raisebox{-0.43ex}{\includegraphics[page=3,height=2.12ex]{figures/platform_icons.pdf}}\hspace{0.31em}Hugging Face}}}
\hypersetup{pdfauthor={Dengzhe Hou},pdftitle={HumanDec: Are System One Decision Models Better Predictors of Human Visual Choices?},pdfsubject={Human visual choice prediction}}
\begin{document}
\maketitle
\begin{abstract}
\upshape
\setlength{\baselineskip}{11.55pt}
Adapting pretrained visual models to human behavior requires choosing an initialization and supervision that capture systematic errors. We ask whether System One decision models offer a transferable advantage over their base counterparts. Using two matched model pairs, we assess (1) starting performance, (2) the effect of human-choice training and (3) transfer from model-vs-human to ImageNet16H. With source-validation temperature calibration, decision checkpoints start ahead in human-choice negative log-likelihood (NLL) in six external settings. Human-choice adaptation reduces or reverses this advantage in \textbf{16 of 18 paired contrasts}, with one opposite and one uncertain shift. Human supervision supplies image-specific information beyond the tested class-average, teacher and shuffled targets. On source-nonoverlapping ImageNet16H, 29 of 36 paired 95\% intervals for overall NLL favor correct-label training, and seven contain zero. On trials where observers err, human-choice training matches their specific wrong choice \textbf{2.2 to 2.7 percentage points more often} across the three source holdouts. These findings motivate evaluating transfer after fine-tuning to guide checkpoint choice and measuring agreement on specific errors alongside overall choice likelihood to assess human supervision.
\par
\end{abstract}

\section{Introduction}
Predicting human visual choices requires explaining errors as well as correct recognition. Consider a degraded bear image that an observer calls a dog (Fig.~\ref{fig:errorcomposition}). A model of that response should assign probability to the specific misidentification. Two models with similar accuracy can make different mistakes. Trial-by-trial error consistency (EC) quantifies agreement beyond chance on which images a model and an observer get right or wrong~\cite{geirhos2020beyond}.

\begin{figure*}[t]
\centering
\includegraphics[width=\textwidth]{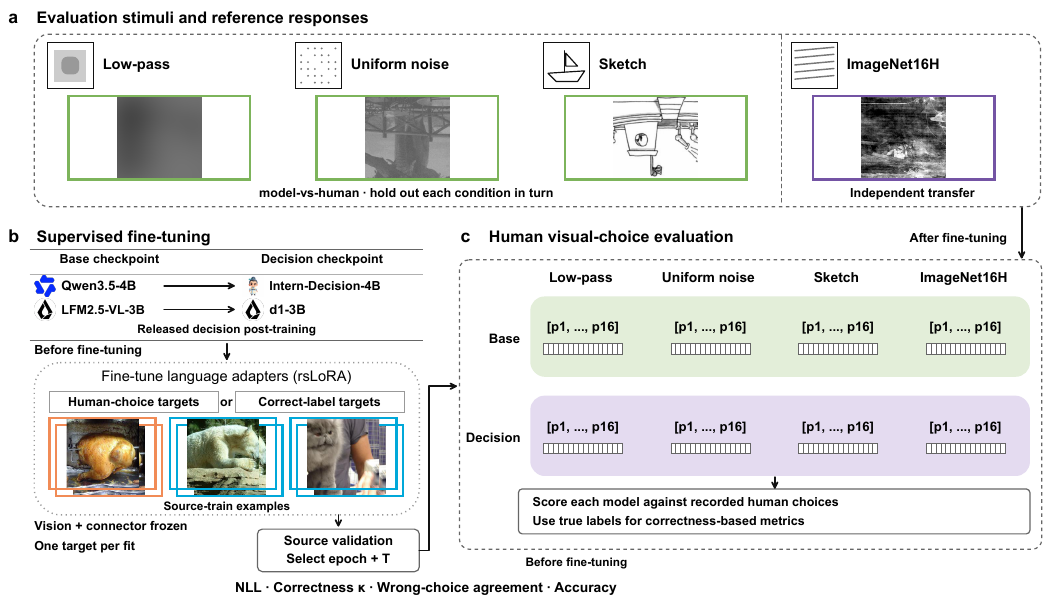}
\caption{Each released checkpoint is adapted independently with human or correct-label targets on source-training images. Source validation selects epochs and temperatures, and the fixed models then predict held-out conditions and independent ImageNet16H choices. Probability vectors are schematic. Correctness $\kappa$ denotes EC. Source images come from model-vs-human~\cite{geirhos2021partial} and transfer images from ImageNet16H~\cite{steyvers2022bayesian}.}
\label{fig:protocol}
\end{figure*}

Released decision checkpoints, including Intern-Decision and d1, expose finite choices and candidate probabilities~\cite{zhu2026intern,liquid2026d1}. Their general-purpose precursors permit paired comparisons of decision post-training. Before adaptation, which checkpoint better predicts human choices in an independent visual experiment? How does that gap change after both are fine-tuned on the same human responses from the source experiment?

Human uncertainty labels and class-level alternatives can affect classifier robustness~\cite{peterson2019human}. Prior work also studies human-judgment fine-tuning of vision-language models and generalization across visual appearances and cognitive domains~\cite{schulzebuschoff2025testing}. Psych-201 reports that post-training can reduce human alignment in language-based behavioral prediction~\cite{binz2026post}.

We compare Qwen3.5-4B with Intern-Decision-4B and LFM2.5-VL-3B with d1-3B under matched image inputs, categorical readouts and adaptation recipes. Source-separated model-vs-human folds hold out low-pass, uniform-noise or sketch images~\cite{geirhos2021partial}. Source-selected checkpoints and temperatures transfer without refitting to ImageNet16H~\cite{steyvers2022bayesian} (Fig.~\ref{fig:protocol}). Alternative targets and visual-input interventions distinguish image-specific human information from class tendencies and target softness.

We make four contributions.
\begin{itemize}
\item \textbf{Starting advantage.} Source-validation temperature calibration yields lower external human-choice NLL for unadapted decision checkpoints in all six related settings.
\item \textbf{Advantage after adaptation.} Human-choice adaptation reduces or reverses this advantage in 16 of 18 paired contrasts, separating starting performance from suitability for adaptation.
\item \textbf{Predicting specific errors across experiments.} Human targets add image-specific information beyond tested alternatives. Correct-label training generally yields lower overall human-choice NLL on ImageNet16H. Human-choice training more often predicts the specific wrong category that an observer selects.
\item \textbf{Reusable evaluation.} We provide the source-separated splits, reference predictions and four-metric scoring as HuDec, supporting matched checkpoint, supervision and transfer tests.
\end{itemize}

\section{Related work}
Human visual alignment requires more than recognition accuracy. Controlled distortions and cue conflicts reveal behavioral differences between humans and models~\cite{geirhos2018generalisation,geirhos2019texture}. Model-vs-human uses EC to compare shared successes and failures~\cite{geirhos2021partial,geirhos2020beyond}. Contrast, depth-cue, contour and MLLM perceptual benchmarks provide complementary diagnostics~\cite{cai2025lowlevel,danier2025depthcues,lonnqvist2025contour,lin2024hvsbench}. CogBattery shows that matching human effects can coexist with different errors and confidence reports~\cite{hou2026cogbattery}. We score human-choice probabilities and wrong-category agreement.

Human uncertainty training and visual-cognition fine-tuning predate this work~\cite{peterson2019human,schulzebuschoff2025testing}. Human-aligned representations, VQA uncertainty and cross-modal choice/attention studies extend this setting~\cite{muttenthaler2025aligning,lan2025uncertainty,hong2026similar}. Behavioral models, post-training audits and mixed generative/discriminative objectives motivate checkpoint comparisons~\cite{binz2025foundation,binz2026post,changortega2026sweetspot}. S1CogBench evaluates System One models on established cognitive tasks~\cite{hou2026s1cogbench}. We compare matched base and decision checkpoints before and after fine-tuning on the same human responses to see how their ability to predict human visual choices changes in an independent experiment.

Visual baselines include CLIP, ResNet and the fixed low-pass control~\cite{radford2021learning,he2016deep,wolff2026lowpass}. Calibration follows validation fitting~\cite{guo2017calibration}, and broader adaptation studies examine robustness under distribution shift~\cite{kumar2022finetuning,wortsman2022robust}. The supplement extends this discussion.

\section{A paired visual-choice protocol}
\subsection{Models and the estimands}
We pair Qwen3.5-4B~\cite{qwen2026qwen35} with Intern-Decision-4B~\cite{zhu2026intern} and LFM2.5-VL-3B~\cite{liquid2026lfm25} with d1-3B~\cite{liquid2026d1}. Base denotes each released precursor, not necessarily a raw pretrained model, and decision its decision-post-trained derivative. Liquid AI initialized d1-3B by averaging LFM2.5-2.6B weights with those of the LFM2.5-VL-3B language backbone before decision fine-tuning~\cite{liquid2026d1release}. Developer-reported decision-task scores are not behavioral results here.

For a fixed model, fold and seed, let $L_{m,t}$ denote human-choice NLL after adaptation from checkpoint $m$ with target $t$. We distinguish the final human-adapted checkpoint difference
\begin{equation}
D_H=L_{\decision,\human}-L_{\base,\human}
\label{eq:dh}
\end{equation}
from the supervision interaction
\begin{equation}
I=(L_{\decision,\human}-L_{\decision,\correct})-
  (L_{\base,\human}-L_{\base,\correct}).
\label{eq:interaction}
\end{equation}
Negative $D_H$ favors the decision checkpoint after human adaptation. Negative $I$ means a larger human-over-correct-label NLL advantage for that checkpoint. Neither quantity is the unadapted checkpoint difference.

\subsection{Visual data and source separation}
Model-vs-human supplies 18,080 stimuli and 84,523 responses in 16 valid categories. Three folds hold out low-pass, uniform-noise or sketch images, using the other 16 datasets for candidate training and validation. Available original ImageNet identifiers group source transformations. Candidates overlapping held-out sources are removed, then training and validation are source-separated. Otherwise, dataset and stimulus identifiers define groups. This separates source IDs without providing universal near-duplicate detection.

Table~\ref{tab:folds} gives source split and external test sizes. The held-out condition is never used for gradient updates, epoch selection or temperature fitting.

\begin{table}[htb]
\centering\footnotesize
\setlength{\tabcolsep}{1.8pt}
\begin{tabular}{@{}lrrrrrrr@{}}
\toprule
 & \multicolumn{5}{c}{\cellcolor{TableSource}\textbf{model-vs-human}} & \multicolumn{2}{c@{}}{\cellcolor{TableExternal}\shortstack{\textbf{ImageNet16H}\\\scriptsize Source-disjoint}} \\
\cmidrule(lr){2-6}\cmidrule(l){7-8}
\textbf{Held-out fold} & \shortstack{\textbf{Train}\\\scriptsize imgs.} & \shortstack{\textbf{Val.}\\\scriptsize imgs.} & \shortstack{\textbf{Test}\\\scriptsize imgs.} & \shortstack{\textbf{Test}\\\scriptsize resp.} & \shortstack{\textbf{Excl.}\\\scriptsize imgs.} & \shortstack{\textbf{Test}\\\scriptsize imgs.} & \shortstack{\textbf{Test}\\\scriptsize resp.} \\
\midrule
Low-pass      & 14,849 & 1,662 & 1,280 & 5,096 (4) & 289 & 3,124 & 18,883 \\
Uniform noise & 14,709 & 1,656 & 1,280 & 5,082 (4) & 435 & 3,160 & 19,098 \\
Sketch        & 15,551 & 1,729 &   800 & 5,560 (7) &   0 & 2,940 & 17,771 \\
\bottomrule
\end{tabular}
\caption{Model-vs-human splits and ImageNet16H test sizes. Source test responses list observer counts in parentheses. Excluded source images share a test original. ImageNet16H excludes originals used in source training or validation. Full ImageNet16H has 4,800 images, 1,200 originals, 28,997 responses and 145 observers.}
\label{tab:folds}
\end{table}

We report both the full ImageNet16H experiment and a source-nonoverlapping view that removes the sources used in each fold's training or validation. Resampling clusters all four phase variants together. Direct transfer retains source-selected epochs and temperatures. In the cross-experiment transfer protocol, ImageNet16H responses support only labeled target-calibration and conditional-scoring diagnostics, never checkpoint selection. Stimuli, observers and procedure change together.

\subsection{Adaptation and supervision controls}
All four models see the images directly, and each pair uses the same prompt and readout. We read the logits of the 16 option symbols at the answer position and normalize over these options. Training shuffles the option order, and evaluation averages the temperature-scaled probabilities from the canonical and reversed orders. We do not generate reasoning or explanations.

We freeze the base weights and the vision encoder and adapt all linear layers of the language model with rank-stabilized LoRA~\cite{hu2022lora,kalajdzievski2023rank}, using rank 16 and alpha 16. AdamW uses learning rate $5\times10^{-5}$ and weight decay 0.01. Every main fit runs three epochs with an effective batch of eight images.

For image $x_i$, let $n_i$ be the number of valid human responses, $h_i$ their empirical distribution and $y_i$ the correct category. The image-item loss is
\begin{equation}
\mathcal{L}_t=-\frac{1}{\sum_i n_i}\sum_i n_i
        \sum_{k=1}^{16}q^{(t)}_{ik}\log p_\theta(k\mid x_i).
\label{eq:loss}
\end{equation}
Human training sets $q^{(H)}_i=h_i$, while correct-label training sets $q^{(G)}_i$ to the one-hot correct label. Their image weights are identical.

Qwen/Intern add three controls (Table~\ref{tab:targets}). These test class, softness and correspondence alternatives, not every soft-label construction.

\begin{table}[htb]
\centering\small
\setlength{\tabcolsep}{3pt}
\begin{tabular}{@{}l>{\raggedright\arraybackslash}p{3.1cm}>{\raggedright\arraybackslash}p{2.3cm}@{}}
\toprule
\rowcolor{TableHeader}\textbf{Target} & \textbf{Training distribution for image $x_i$} & \textbf{What it tests} \\
\midrule
\multicolumn{3}{@{}l@{}}{\textbf{Matched supervision (all four checkpoints)}} \\
\cellcolor{TableHuman!17}\textbf{Human (H)} & its own human choices $h_i$ & image-specific human targets \\
\cellcolor{TableCorrect!14}Correct (G) & one-hot correct category & correct-category supervision \\
\midrule
\multicolumn{3}{@{}l@{}}{\textbf{Additional controls (Qwen/Intern)}} \\
Class average & pooled human choices of its true class & class-level confusion only \\
Distillation & mean prediction of three correct-trained models & soft labels from a model \\
Shuffled & human choices $h_j$ of another image in its stratum & image-response correspondence \\
\bottomrule
\end{tabular}
\caption{Training targets. All targets share images, image weights $n_i$ and optimization. Shuffling stays within strata of true class, dataset, condition and response count.}
\label{tab:targets}
\end{table}

Qwen and Intern have 90 fits over five targets, three folds and three seeds, and LFM2 and d1 have 36 fits with human and correct targets only. Epochs are chosen on source validation under two rules. The original rule selects correct-label fits by correct-label validation NLL and all other fits by human validation NLL. Common selection uses human validation NLL for every target, which changes the epoch of 21 of the 36 correct-label fits. Evaluated models always use temperatures fitted on source human validation.

As a strong visual baseline, we score zero-shot OpenCLIP ViT-H/14 (H14) with LAION-2B weights and 80 prompts over the 16 categories, both on the original images and after a fixed Gaussian blur with $\sigma=2.5$ pixels~\cite{wolff2026lowpass}. Its temperature is fitted on source human validation in each fold.

Two diagnostics reuse saved predictions. Target calibration refits the temperature, alone or with 15 category biases, on 20\% of ImageNet16H sources and scores the remaining 80\%. Accuracy-matched comparisons compare source and external tests on images with the same observed human accuracy. The supplement gives details.

\subsection{Metrics and paired uncertainty}
The primary metric is human-choice NLL averaged over all valid trials,
\begin{equation}
\nll=-\frac{1}{N}\sum_{s,i}\log p(r_{si}\mid x_i),
\end{equation}
where $r_{si}$ is observer $s$'s response to image $x_i$. We also report three secondary metrics. We compute EC as Cohen's $\kappa$ on the model's and each observer's correct or incorrect responses to the same images. Wrong-category agreement is the fraction of jointly wrong trials on which both choose the same wrong category. Accuracy weights each image equally.

Each contrast compares two models on the same 2,000 bootstrap draws of observers and images, and we report a 95\% percentile interval for each seed. These intervals reflect the sampled images and observers, not retraining with another seed. Counts of favorable intervals summarize related configurations rather than independent replications, and intervals that contain zero do not show equivalence. The supplement gives full definitions, and an accompanying numerical archive lists every model, seed and metric.

We also measure top-choice agreement on human-error trials. A model matches an observer if its most probable of all 16 categories equals the observer's wrong choice. We weight original images equally, displayed variants equally within each original, and responses equally within each variant, then renormalize over human-error trials. Paired 95\% intervals use 2,000 resamples of original images and exponential weights shared by responses from the same observer. Original images shared between tests are resampled jointly, and observer weights are drawn independently in each study.

\paragraph{Additional checks.}
Two further checks use the same metrics. The first changes the visual input of every checkpoint under common selection, before and after adaptation, either by showing another image of the same category and condition or by zeroing the image tokens, while keeping the original human targets and temperatures. The second retrains on the low-pass fold, mostly with seed 0, for 10 epochs or with a trainable visual encoder or connector. The supplement describes both.

\section{What decision post-training transfers}
\subsection{The starting advantage is conditional}
\label{sec:startadvantage}

\begin{figure*}[t]
\centering
\includegraphics[width=\textwidth]{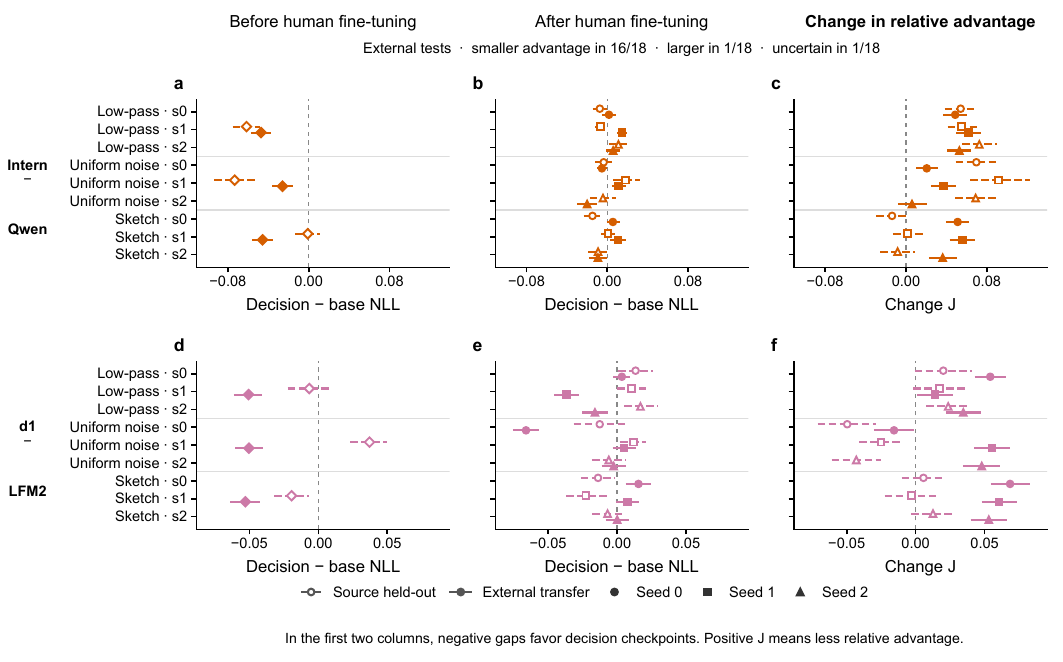}
\caption{Decision-minus-base human-choice NLL before and after human adaptation and their paired change $J$. Rows show Intern/Qwen and d1/LFM2. Open dashed marks show source tests, filled marks source-nonoverlapping ImageNet16H, and diamonds the shared unadapted gap. Negative gaps favor the decision checkpoint, and positive $J$ means a smaller relative advantage. Bars are paired 95\% intervals from 2,000 draws.}
\label{fig:decisiontrajectory}
\end{figure*}

With source-validation temperature calibration, both released decision checkpoints predict external human choices better than their bases before any adaptation. On source-nonoverlapping ImageNet16H, NLL favors the decision checkpoint in all six comparisons of two pairs and three folds (Fig.~\ref{fig:decisiontrajectory}).

Human adaptation shrinks this advantage. We measure the change as $J=D_H-D_0$, where $D_0$ is the decision-minus-base NLL before adaptation and $D_H$ is Eq.~\ref{eq:dh}, so positive $J$ means that the decision checkpoint loses ground. Of 18 paired intervals over two pairs, three folds and three seeds, 16 are positive, one is negative and one contains zero. After adaptation, the decision checkpoint is better in six intervals, the base in four, and eight contain zero.

Losing ground does not always mean getting worse. Qwen improves in all nine external settings and Intern in six, with the other three containing zero, so for this pair the shift reflects a smaller gain for Intern. For low-pass seed 0, Qwen improves by 0.0816 and Intern by 0.0328, giving $J=0.0487$ $[0.0367,0.0605]$. In the other pair, d1 gets worse in seven of nine settings, while LFM2 improves in one, gets worse in three and has five zero-containing intervals.

The starting advantage also depends on calibration and input. Without calibration, Intern is worse than Qwen in all three folds, whereas d1 remains better than LFM2. With calibration but without visual content, Intern keeps all three advantages, while all three d1 gaps contain zero.

The supervision interaction $I$ (Eq.~\ref{eq:interaction}) has no consistent sign. Under common selection, source intervals are three favorable, two adverse and 13 zero-containing, and external intervals are seven favorable, five adverse and six zero-containing. Neither selection rule shows that decision checkpoints are a consistently better starting point for human adaptation, or that decision post-training is ineffective.

\subsection{Human labels supply image-specific information}
\begin{figure*}[t]
\centering
\includegraphics[width=\textwidth]{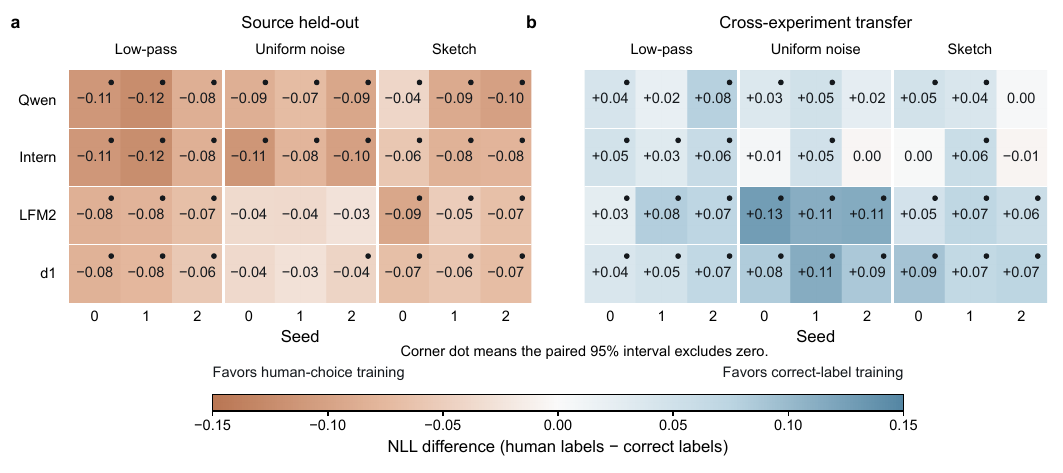}
\caption{Human-minus-correct-label NLL under common selection for every model, held-out fold and seed, on source tests and source-nonoverlapping ImageNet16H. Orange favors human supervision and blue favors correct labels. Dots mark paired 95\% intervals that exclude zero.}
\label{fig:humantransfer}
\end{figure*}

Human training predicts held-out human choices better than correct-label training. Under the original selection rule, its NLL is lower in all 36 model, fold and seed configurations, and the paired interval is below zero in 32. These are nine of nine for Qwen and Intern, six of nine for LFM and eight of nine for d1, and the other four intervals contain zero. Common selection gives 31 favorable and five zero-containing intervals (Table~\ref{tab:selectioncontrols} and Fig.~\ref{fig:humantransfer}).

\begin{table}[htb]
\centering\small
\setlength{\tabcolsep}{3pt}
\begin{tabular}{@{}lrrr@{}}
\toprule
\rowcolor{TableHeader}\textbf{Comparison} & \shortstack{\textbf{Favors}\\\textbf{human}} & \shortstack{\textbf{CI contains}\\\textbf{zero}} & \shortstack{\textbf{Favors}\\\textbf{comparator}} \\
\midrule
\rowcolor{TableSource!70}\multicolumn{4}{@{}l@{}}{\textbf{model-vs-human}} \\
Original H vs G & \textbf{32} & 4 & 0 \\
Common H vs G & \textbf{31} & 5 & 0 \\
\midrule
\rowcolor{TableExternal!70}\multicolumn{4}{@{}l@{}}{\textbf{ImageNet16H (source-nonoverlapping)}} \\
Original H vs G & 0 & 9 & \textbf{27} \\
Common H vs G & 0 & 7 & \textbf{29} \\
H vs plain H14 & 0 & 7 & \textbf{29} \\
H vs low-pass H14 & \textbf{36} & 0 & 0 \\
\bottomrule
\end{tabular}

\caption{Counts of paired 95\% NLL intervals over 36 configurations per row. Bold marks a direction supported by more than half the intervals. H and G denote human and correct-label targets. The OpenCLIP ViT-H/14 rows use common-selection human models.}
\label{tab:selectioncontrols}
\end{table}

The gain lies in the predicted choices rather than in recognition. Under original selection, human training improves EC in 19 of 36 cases and wrong-category agreement in 17, with the remaining intervals containing zero. Accuracy favors correct labels in 22 cases and human targets in two, with 12 containing zero.

To see where the NLL gain comes from, we split the cross-entropy exactly into a correctness term and an error-content term. Write $a_i=h_i(y_i)$, $b_i=p_i(y_i)$, and let $\widetilde h_i$ and $\widetilde p_i$ be distributions renormalized over the wrong categories. Then
\begin{equation}
\mathrm{CE}(h_i,p_i)=\mathrm{CE}_{\mathrm{Bern}}(a_i,b_i)
+(1-a_i)\mathrm{CE}(\widetilde h_i,\widetilde p_i).
\label{eq:decomposition}
\end{equation}
The second term measures which wrong categories humans choose, weighted by how often they err. Human training lowers this term in all 36 intervals, with point reductions from 0.0300 to 0.1291, and the conditional wrong-category NLL also favors human targets in all 36. Human supervision therefore improves the predicted content of human errors, even in the four configurations where total NLL does not separate under the original rule.

\begin{figure*}[t]
\centering
\includegraphics[width=\textwidth]{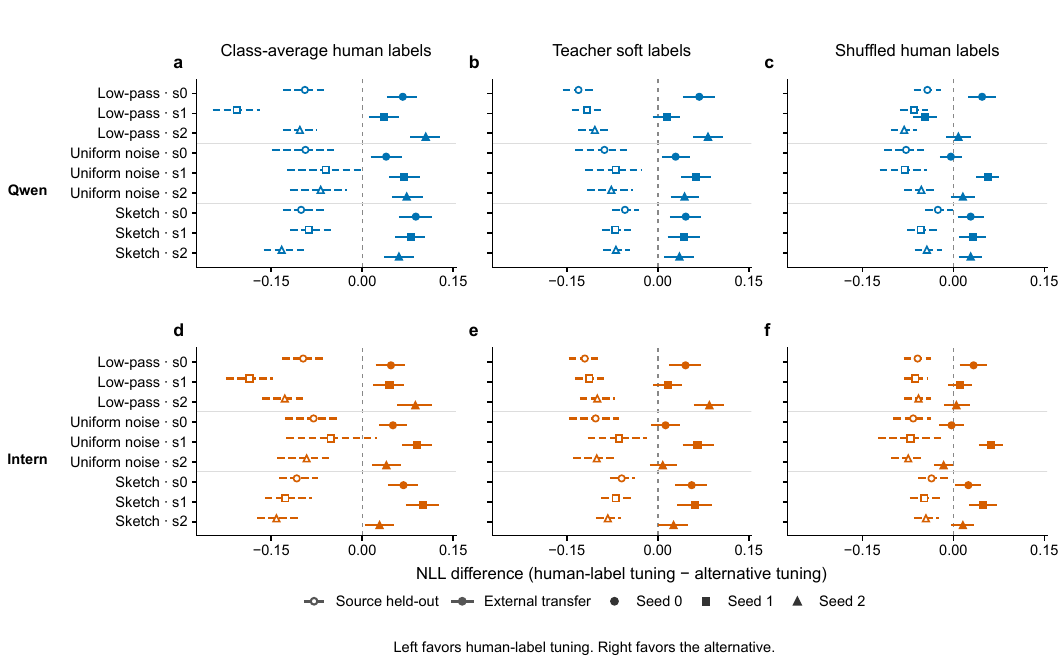}
\caption{Human-minus-alternative NLL for Qwen and Intern against class-average, distillation and conditionally shuffled targets, for every held-out fold and seed. Open dashed marks show source tests and filled marks source-nonoverlapping ImageNet16H. Negative values favor human targets. Bars are paired 95\% intervals from 2,000 draws.}
\label{fig:targetcontrols}
\end{figure*}

On the source tests, the Qwen/Intern controls show that this benefit depends on the responses to each specific image (Fig.~\ref{fig:targetcontrols}). Human targets beat distillation targets in all 18 paired NLL intervals and conditionally shuffled targets in all 18, and they beat class averages in 17 of 18, with one interval containing zero. Both the base and the decision checkpoint show this benefit, so it does not establish a contribution unique to decision post-training. On source-nonoverlapping ImageNet16H, the alternatives do better. Human targets are worse than class averages in all 18 intervals and worse than distillation targets in 14, with four containing zero. Against shuffled targets, two intervals favor human targets, seven contain zero and nine are adverse.

\subsection{Image dependence and checkpoint advantage are distinct}
The two input checks described in the protocol answer two separate questions. Does the human-supervision benefit depend on the specific image? And does the decision checkpoint lose its advantage because of what it sees?

The human-supervision benefit depends on the specific image. When each image is replaced by another image of the same category and condition, the human-minus-correct-label NLL difference becomes less favorable to human training in all 36 source intervals. On source-nonoverlapping ImageNet16H, it becomes less favorable in 14 intervals, and 22 contain zero.

Swapping the image does not detectably change the decision checkpoint's loss of advantage. For each input view $v$, we recompute the adaptation shift $J_v=D_{H,v}-D_{0,v}$ and compare it with the original $J$. On source-nonoverlapping ImageNet16H, all \VisualSwapExternalZero{} of 18 same-category replacement intervals for $J_v-J_{\rm original}$ contain zero. Removing visual content instead changes $J$ in both directions, with \VisualMaskExternalNegative{} negative, \VisualMaskExternalPositive{} positive and \VisualMaskExternalZero{} zero-containing intervals. Human supervision therefore adds image-specific information, whereas swapping the image leaves no detectable trace on the relative loss of the decision checkpoint, although removing visual content does change it.

\begin{figure*}[!t]
\centering
\includegraphics[width=\textwidth]{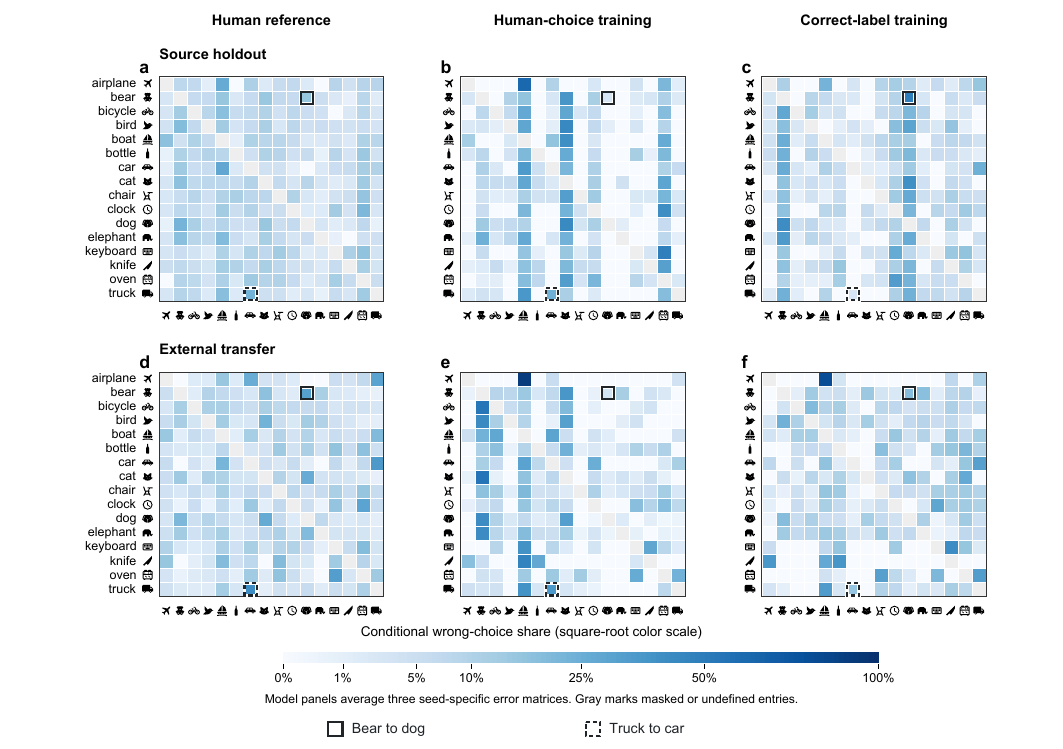}
\caption{Conditional wrong choices for Qwen3.5-4B with low-pass held out, on the source holdout (top) and source-nonoverlapping ImageNet16H (bottom). Each panel averages three seed matrices of trial-weighted wrong choices, normalized within each true class (rows) over chosen classes (columns). Gray marks correct or undefined entries. Solid and dashed outlines mark the same two category pairs in all panels (bear to dog and truck to car, respectively). The outlines are reading guides.}
\label{fig:errorcomposition}
\end{figure*}

\subsection{Transfer of overall choice likelihood and specific wrong choices}
On source-nonoverlapping ImageNet16H, 29 of 36 paired overall NLL intervals favor correct-label training under common selection, and seven contain zero (Fig.~\ref{fig:humantransfer}). Among trials with an observer error, human-choice training matches the specific wrong category 2.15 to 2.71 percentage points more often across the three source holdouts, with all three intervals favoring it (Table~\ref{tab:human-error-choices}).

Under original selection, 27 of 36 source-nonoverlapping NLL intervals favor correct-label training and nine contain zero. LFM and d1 do worse with human training in all nine settings, Qwen in six and Intern in three. On full ImageNet16H, 28 intervals favor correct-label training and eight contain zero. In the source tests on human-error trials, the low-pass and uniform-noise intervals favor human-choice training, and the sketch interval includes zero.

\begin{table}[t]
\centering
\small
\setlength{\tabcolsep}{3pt}
\begin{tabular}{@{}lrr@{}}
\toprule
\rowcolor{TableHeader}Source holdout & \cellcolor{TableSource}model-vs-human & \cellcolor{TableExternal}ImageNet16H \\
\midrule
Low-pass & \(+5.44\;[3.27,7.67]\) & \(+2.67\;[1.24,4.15]\) \\
Uniform noise & \(+4.48\;[1.59,8.75]\) & \(+2.15\;[0.61,3.77]\) \\
Sketch & \(+1.95\;[-1.42,4.70]\) & \(+2.71\;[1.12,4.20]\) \\
\bottomrule
\end{tabular}
\caption{Top-choice agreement on trials with an observer error. Positive differences favor human-choice training, in percentage points. Values average four checkpoints and three seeds equally. Brackets give paired 95\% intervals. ImageNet16H columns use the corresponding source-nonoverlapping views.}
\label{tab:human-error-choices}
\end{table}

Figure~\ref{fig:errorcomposition} summarizes the conditional wrong-choice patterns for the Qwen low-pass comparison, averaging the three separately normalized seed matrices. The supplement retains each seed separately.

This does not mean that adaptation fails externally. Human training still improves Qwen in all nine external settings and Intern in six (Sec.~\ref{sec:startadvantage}), and after human training the decision checkpoint still beats its base in six of 18 external intervals. Under original selection, the supervision interaction $I$ favors the decision checkpoint in seven external intervals, two for Intern and five for d1, and is adverse in two.

NLL gains on source tests and external NLL advantages are separate questions. A checkpoint can gain more from human supervision on source NLL without an external NLL advantage, and it can hold an external NLL advantage without transferring the human-over-correct-label NLL benefit better. Both matter when choosing a starting point for adaptation.

\subsection{Selection rule, visual baseline and recalibration}

Common selection leaves the total-NLL pattern unchanged (Table~\ref{tab:selectioncontrols}) but changes the error-content term. That term favors human training in all \CommonInternalWrongFav{} source comparisons and in only five external ones, with 31 containing zero, against 16 favorable external intervals under the original rule. The external correctness term is adverse in 32 intervals.

The strong visual baseline separates the two settings. On source tests, human training beats zero-shot OpenCLIP H14 in all 36 NLL intervals and its low-pass variant in 29, with seven containing zero. On source-nonoverlapping ImageNet16H, plain H14 beats human training in 29 comparisons, with seven containing zero, whereas low-pass H14 loses all 36. These comparisons use our splits and calibration and are not comparable to the published aggregate EC ranking.

Recalibrating on the target data does not consistently restore the human-supervision benefit. On the shared 80\% source-nonoverlapping test subsets, a temperature with category biases fitted on the other 20\% improves every common-selection model in all 36 paired NLL contrasts, but human-minus-correct NLL stays adverse in 35 intervals, with one containing zero. A temperature alone leaves all 36 adverse.

Comparing images with the same observed human accuracy gives mixed results. Where all observers erred, human training is better in 35 source intervals, but on full ImageNet16H it is better in five and uncertain in 31, and on the source-nonoverlapping view all 36 intervals contain zero although 31 point estimates favor it. Where all observers were correct, the source-nonoverlapping view has three favorable, 23 adverse and ten zero-containing intervals.

\subsection{Sensitivity to adaptation settings}
Longer training does not consistently restore a decision-checkpoint advantage. With frozen vision and human targets, external Intern-minus-Qwen NLL after 10 epochs is $-0.0737$ $[-0.0885,-0.0584]$ at seed 0, $0.0350$ $[0.0203,0.0488]$ at seed 1 and $-0.0030$ $[-0.0157,0.0094]$ at seed 2. Human targets still beat correct labels on the source test in all 12 frozen Qwen/Intern comparisons, but none of the external differences favors them, with seven adverse and five containing zero. Unfreezing the visual encoder of human-trained models gives five adverse and three zero-containing external intervals across eight comparisons, and training the connector leaves three of four external human-minus-correct intervals adverse. Table~\ref{tab:training-settings} summarizes these checks.

\begin{table*}[!b]
\centering\small
\setlength{\tabcolsep}{4pt}
\begin{tabular}{@{}p{.28\textwidth}>{\centering\arraybackslash}p{.05\textwidth}>{\centering\arraybackslash}p{.145\textwidth}>{\centering\arraybackslash}p{.145\textwidth}>{\centering\arraybackslash}p{.145\textwidth}>{\centering\arraybackslash}p{.145\textwidth}@{}}
\toprule
\rowcolor{TableHeader}\textbf{Setting change} & \textbf{$n$} & \multicolumn{2}{c}{\cellcolor{TableHuman!17}\textbf{Human targets}} & \multicolumn{2}{c}{\cellcolor{TableCorrect!14}\textbf{Correct targets}}\\
\cmidrule(lr){3-4}\cmidrule(l){5-6}
\rowcolor{TableHeader}\multicolumn{2}{@{}l}{\footnotesize Improved / CI includes zero / Worse} & \cellcolor{TableSource!75}\textbf{Source} & \cellcolor{TableExternal}\textbf{External} & \cellcolor{TableSource!75}\textbf{Source} & \cellcolor{TableExternal!65}\textbf{External}\\
\midrule
10 vs 3 epochs, frozen vision & 8 & 0 / 4 / 4 & \textbf{4 / 1 / 3} & 2 / 6 / 0 & 1 / 3 / 4\\
10 vs 3 epochs, unfrozen vision & 4 & 0 / 1 / 3 & \textbf{0 / 2 / 2} & 1 / 2 / 1 & 1 / 1 / 2\\
Unfrozen vs frozen vision & 8 & 0 / 4 / 4 & \textbf{0 / 3 / 5} & 0 / 1 / 7 & 1 / 3 / 4\\
Updated vs fixed connector & 4 & 0 / 2 / 2 & \textbf{2 / 0 / 2} & 2 / 2 / 0 & 2 / 2 / 0\\
\bottomrule
\end{tabular}

\caption{Low-pass training-setting changes. Each cell counts the direction of the paired 95\% NLL intervals across $n$ comparisons. Bold highlights human-target transfer. External results use source-nonoverlapping ImageNet16H. The first row uses Qwen/Intern seeds 0--2 and LFM2/d1 seed 0. The remaining rows use seed 0.}
\label{tab:training-settings}
\end{table*}

\enlargethispage{-12pt}
\AddToHookNext{para/end}{\baselineskip=12pt\relax}
Observer history adds a small benefit. When the input includes the same observer's four preceding image--response trials, NLL is 1.4672, against 1.4773 without history and 1.4815 with another observer's responses to the same images. Own minus no history is $-0.0101$ $[-0.0195,-0.0009]$ and own minus other history is $-0.0143$ $[-0.0212,-0.0072]$, while all secondary-metric intervals contain zero. An earlier inference-only version without history training was worse, by $0.6435$ NLL for one observer. The supplement gives the design.\par

\section{Discussion and limits}
Binz et al.~\cite{binz2026post} show that common post-training objectives can reduce behavioral alignment in experiments presented as text. We examine what happens when released decision checkpoints and their base counterparts are subsequently fine-tuned on the same human visual choices. In our two model pairs, the decision checkpoints lead after calibration, but this gap usually narrows after fine-tuning. Which checkpoint predicts human choices better at the start can therefore differ from which performs better after adaptation.

Schulze Buschoff et al.~\cite{schulzebuschoff2025testing} found that gains from human-judgment fine-tuning can carry over to a new visual appearance but vary across tasks. Cai et al.~\cite{cai2025lowlevel} similarly found that agreement with humans varies across properties of low-level vision. We find a distinction within a single recognition task across two experiments. On ImageNet16H, correct-label training generally yields lower NLL across all human responses, while training on human choices more often predicts the exact category an observer selects when they misidentify an image.

Two observations constrain why the calibrated head start fades. Intern keeps its advantage when visual content is removed, so part of it reflects output tendencies rather than visual processing, and human adaptation reshapes exactly these output preferences. d1 loses its advantage without vision, yet its external fit worsens after adaptation in seven of nine settings. Calibration also moves $J$ in opposite directions for the two pairs, upward in all nine Qwen/Intern intervals and downward in all nine LFM2/d1 intervals, and without calibration $J$ is positive in only nine of 18 intervals. The two decision checkpoints therefore lose ground for different reasons, which suggests evaluating decision post-training separately for its output prior and for its visual evidence.

Three limits matter most. The external test changes images, observers and procedure at once, so the transfer result cannot be traced to a single change. A study using identical visual stimuli found different patterns of frontal-midline theta activity before self-initiated and instructed shifts of attention~\cite{hou2025attention}. The low-pass and uniform-noise test sets have only two to four human responses per image, against six or seven in ImageNet16H. All fits use one adaptation recipe on one 16-class recognition task, so other recipes or tasks could change the size of the effects. Future work should test whether these patterns hold for other decision post-training recipes, larger models and visual tasks beyond 16-class recognition. The supplement lists further limits.

\section{Conclusion}
\textbf{Are System One decision models better predictors of human visual choices? Before adaptation and with temperature calibration, yes. Once both members of a pair learn the same human choices, the gap largely closes}, with the decision checkpoint ahead in six of 18 paired comparisons, the base in four and eight indistinguishable. The advantage shrinks or reverses in 16 of 18 comparisons, so a better calibrated start does not identify a better starting point for adaptation. Human labels behave differently. \textbf{On the source tests they add information specific to each image}, beyond class-level, distilled and shuffled targets. On ImageNet16H, training on human choices matched observers' specific wrong choices 2.2 to 2.7 percentage points more often across the three source holdouts, even though correct-label training generally achieved lower overall choice NLL. HumanDec provides a protocol for comparing checkpoints after matched fine-tuning and testing them in an independent experiment. Evaluations of human visual behavior should report both overall choice likelihood and whether the model predicts the specific mistakes people make.

\label{page:bodyend}
\section*{LLM Usage Statement}
OpenAI Codex/ChatGPT and Anthropic Claude assisted with research implementation, analysis and manuscript preparation. The author takes responsibility for all content.
\section*{Data availability}
Behavioral data are available from the original \href{https://github.com/bethgelab/model-vs-human/tree/79ed7cd1d9ac3a114e71b6638f0a98bdd627316c/raw-data}{model-vs-human repository}~\cite{geirhos2021partial} and the \href{https://osf.io/2ntrf/}{ImageNet16H OSF project}~\cite{steyvers2022bayesian}.

{\small\setlength{\bibsep}{0pt}\fontsize{9pt}{10.5pt}\selectfont\bibliographystyle{ieeenat_fullname}\bibliography{references}}

\appendix
\section*{Appendix}
\section{Evaluation details}
\paragraph{Prompt and readout.}
The canonical category order is airplane, bear, bicycle, bird, boat, bottle, car, cat, chair, clock, dog, elephant, keyboard, knife, oven and truck. The prompt asks which category the image shows, and the readout normalizes the probabilities of the 16 legal option symbols. Training uses random category orders. Evaluation uses the canonical and reversed orders and averages the two probability vectors after mapping them to the same categories.

\paragraph{Calibration and epoch selection.}
Temperatures are chosen on a 61-point log-spaced grid from 0.25 to 8 by minimizing the source-validation human-choice NLL of the averaged temperature-scaled probabilities. The temperature then stays fixed on the held-out fold and in external transfer. Under the original rule, correct-label fits select epochs by correct-label validation NLL and all other targets by human validation NLL. Common selection applies human validation NLL to every target. All evaluated models use temperatures fitted on human validation.

\paragraph{Control targets.}
Class-average targets use training responses only. Conditional shuffling applies a fixed cyclic derangement within each stratum that contains more than one image, and single-image strata keep their own target. The same shuffled targets serve both checkpoints and all three training seeds. Distillation targets average the training-image predictions of the model's three existing correct-label fits over both option orders, each at its own correct-label-validation epoch and temperature.

\paragraph{Optimization.}
The learning-rate multiplier at scheduler step $s$ is $\min(1,(s+1)/100)\max(0,1-s/S)$, with $S=3\lceil M/8\rceil$ for $M$ training images, so warm-up and linear decay are multiplied rather than applied in sequence. The implementation applies a constant rescaling to the count-weighted objective of the main paper.

\paragraph{Baselines.}
The cheap comparator combines temperature scaling with a source-human response prior fitted on source validation, and the prior-only baseline makes no image-dependent prediction. Frozen CLIP and ResNet-50 use their recorded preprocessing and label adapters. The fixed Gaussian OpenCLIP variant applies a $21\times21$ kernel to the original RGB image before preprocessing.

\paragraph{Data.}
Original ImageNet identifiers group different distortions of one source, and dataset and stimulus identifiers are used where no source is available. We cannot rule out that model pretraining included these images. ImageNet16H contains 4,800 images from 1,200 sources with 28,997 responses from 145 observers. For each source fold, the source-nonoverlapping view removes the sources used in that fold's training or validation. This leaves 3,124 images and 18,883 responses for the low-pass fold, 3,160 and 19,098 for uniform noise, and 2,940 and 17,771 for sketch.

\paragraph{Scoring and uncertainty.}
NLL weights valid human trials equally, and the model's choice is the most probable legal option after order averaging. EC is computed within observer--condition groups as $\kappa=(p_o-p_e)/(1-p_e)$, where $p_o$ is the observed correctness agreement and $p_e$ the agreement expected from the two correctness rates. Wrong-category agreement divides the trials on which both are wrong and choose the same wrong category by the trials on which both are wrong, so it can be imprecise when joint errors are few. Accuracy weights images equally. All compared models share the 2,000 bootstrap draws, which resample observers and stimuli within condition and resample ImageNet16H by original source with its four phase variants. Undefined estimates are reported as missing rather than zero, and invalid draws are never dropped. All NLL differences are computed within one evaluation protocol, never across the raw NLL scales of different datasets.

\paragraph{Additional results in the archive.}
Calibration moves $J$ in opposite directions for the two pairs, upward in all nine Qwen/Intern intervals and downward in all nine LFM2/d1 intervals. Without calibration, $J$ is positive in nine of the 18 intervals and negative in the other nine. Original epoch selection also gives favorable, adverse and zero-containing checkpoint differences and supervision interactions. The numerical archive reports these results, both terms of the cross-entropy decomposition for all seeds, all four-metric contrasts and individual arm scores. It also keeps the source-validation learning curves, which are not held-out test curves.

\begin{figure*}[!t]
\centering
\includegraphics[width=\textwidth]{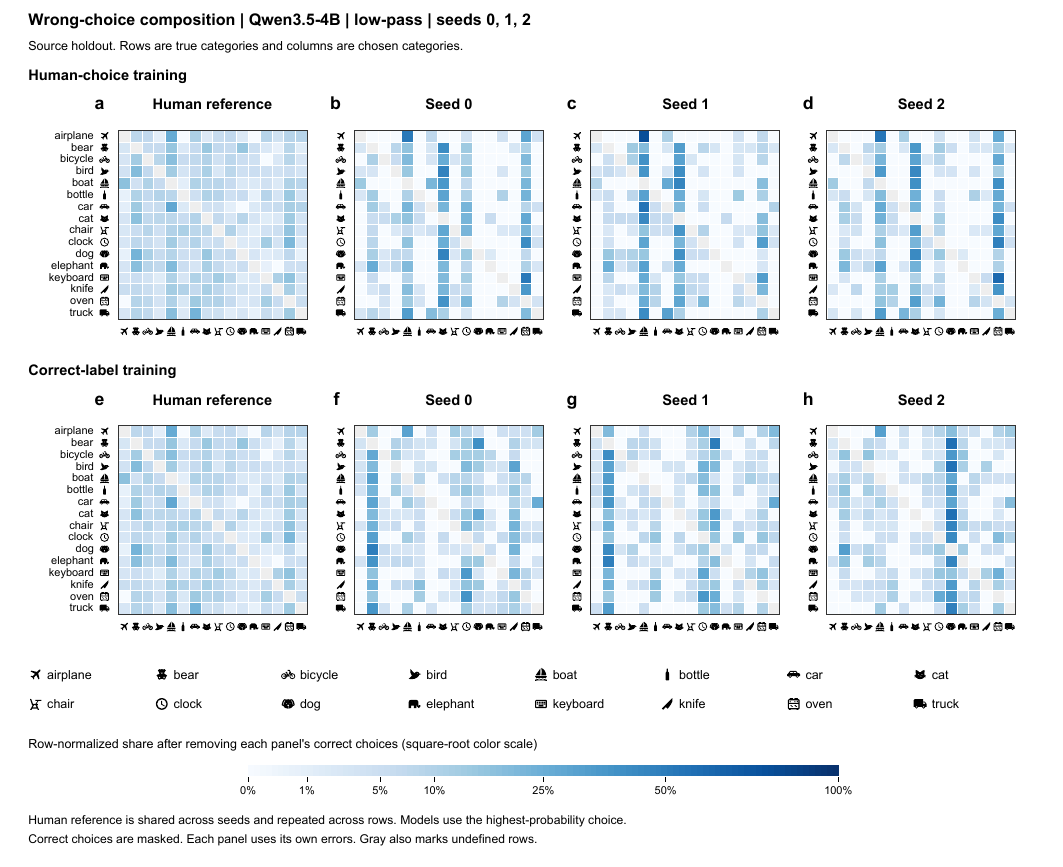}
\caption{Wrong-choice matrices on held-out low-pass images for Qwen3.5-4B. Rows compare human-choice and correct-label training. Seeds 0, 1 and 2 appear separately beside the shared human reference. Entries are trial-weighted frequencies of the highest-probability choice, normalized within the true class after removing each panel's correct choices. The square-root color scale displays percentages. Gray marks masked correct choices or undefined entries. Complete matrices for every model, holdout and seed remain in the numerical archive.}
\label{fig:confusionsource}
\end{figure*}

\begin{figure*}[!t]
\centering
\includegraphics[width=\textwidth]{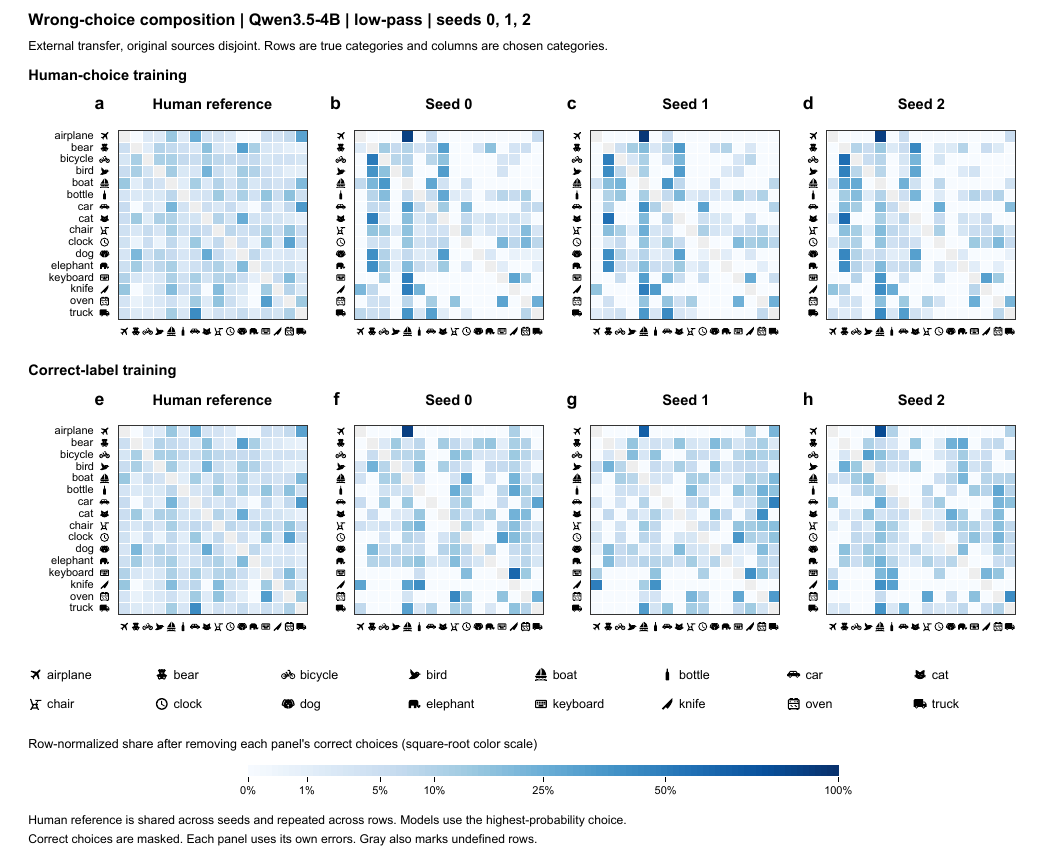}
\caption{Wrong-choice matrices on source-nonoverlapping ImageNet16H for Qwen3.5-4B adapted with low-pass held out. Rows compare human-choice and correct-label training. Seeds 0, 1 and 2 appear separately beside the shared human reference. Entries are trial-weighted frequencies of the highest-probability choice, normalized within the true class after removing each panel's correct choices. The square-root color scale displays percentages. Gray marks masked correct choices or undefined entries. Complete matrices and all four metrics remain in the numerical archive.}
\label{fig:confusionexternal}
\end{figure*}

\paragraph{Error-category matrices.}
\label{sec:confusionmatrices}
Figures~\ref{fig:confusionsource} and~\ref{fig:confusionexternal} show which wrong categories Qwen3.5-4B chooses after human-choice and correct-label training with low-pass held out, next to the human reference, on source and external tests. They give the individual-seed detail behind the mean matrices in the main paper and summarize conditional frequencies of wrong argmax choices.

\section{Additional diagnostic methods}
\paragraph{Visual-input interventions.}
Same-category substitution shows each image's donor, a different source with the same correct category, condition and $224\times224$ raw size, assigned by a fixed cyclic map chosen without human responses. Scoring still uses the original image's human responses, so accuracy is unchanged by construction while the image-to-response correspondence is broken. Full removal zeros every visual token at the language-model input and keeps the text, placeholders, token count, attention and positions. Removal can create unfamiliar inputs, so its effect may include output tendencies under the fixed calibration. Neither check changes model parameters or temperatures, and the bootstrap conditions on the fixed donor map.

\paragraph{Target-label calibration.}
A fixed split assigns 20\% of the original ImageNet16H sources within each correct category to calibration and 80\% to testing, keeping all four phase variants together. The full view has 960 calibration images with 5,798 responses and 3,840 test images with 23,199 responses, and each source-nonoverlapping view intersects this split with its eligible sources. For cached order-averaged probabilities $p_i$, the transform is
\begin{equation}
p_i'=\operatorname{softmax}(\alpha\log p_i+b).
\end{equation}
Temperature-only fitting sets $b=0$, and the second transform fits 15 biases with $\sum_kb_k=0$. Parameters minimize calibration human-choice NLL with L-BFGS-B and $\alpha\in[0.125,4]$, starting from the identity. The principle follows Guo et al.~\cite{guo2017calibration}, but the transform acts on averaged probabilities and human responses rather than correct-label logits. Intervals resample test sources and observers and condition on the fitted parameters. Because ImageNet16H had already been analyzed in aggregate, these are diagnostics rather than an unseen benchmark.

\paragraph{Observed-human-accuracy comparisons.}
Coarse groups use observed correct-response fractions in $[0,.25)$, $[.25,.5)$, $[.5,.75)$ and $[.75,1]$, and exact-fraction scoring keeps every fraction shared by source and external data. These comparisons condition on observed responses and do not equate latent difficulty, image composition or observer precision. Human accuracy is 78.35\% on ImageNet16H, against image-equal training accuracies of 65.16\%, 65.06\% and 62.84\% in the three source folds.

\paragraph{Training duration and visual updates.}
On the low-pass fold with seed 0, we cross the four checkpoints and two targets with frozen or trainable visual encoders and independent 3- or 10-epoch schedules, giving 32 configurations. Sixteen further Qwen/Intern frozen-vision fits repeat both targets and schedules at seeds 1 and 2. Trainable encoders use learning rate $10^{-5}$ with the connector frozen. Eight more fits train only the connector at $10^{-5}$ with frozen vision for three epochs. All other settings match the main fits, and all selections use source human validation. The 10-epoch schedule starts from the original checkpoint with its own decay horizon, so it is not a continuation of the 3-epoch run. These source-selected schedules do not establish convergence or a best training recipe. Connector updates improve human-target external NLL for Qwen and Intern but worsen it for LFM2 and d1. Freezing the visual encoder in the main fits only defines which parameters are trained. It does not imply that predictions use text alone or that human-like errors arise in the language model.

\section{Within-ImageNet16H adaptation}
\label{sec:within16h}
\paragraph{Question and split.}
This control asks whether human-choice supervision helps when training and testing remain within ImageNet16H, in contrast to transfer from model-vs-human. A class-stratified split assigns the 1,200 original sources to 960 training, 120 validation and 120 test sources using split seed 20261117. All four phase variants stay with their original source, giving 3,840 training, 480 validation and 480 test images. The test set contains 2,904 responses from 145 observers. Observers may occur in more than one image split, so this control measures generalization to held-out original sources within one experiment rather than to new observers or a new procedure. ImageNet16H was already used in the transfer analyses before this split was constructed.

\paragraph{Training and selection.}\leavevmode\newline
The four checkpoints are each trained with human-choice and correct-label targets at seeds 0, 1 and 2, giving 24 complete fits. All use the same three-epoch language-adapter recipe as the main experiment, with frozen visual encoders and connectors, rank 16, alpha 16, learning rate $5\times10^{-5}$, weight decay 0.01 and effective batch size eight. Each fit completes 1,440 updates. Every epoch and temperature is selected using human-choice NLL on the within-dataset validation split, including for correct-label training. Four unadapted checkpoints are calibrated on that same validation split. No test responses enter training, selection or calibration. Unlike direct cross-experiment transfer, this control deliberately fits to ImageNet16H training and validation responses. Table~\ref{tab:within16h-arms} lists every trained arm and its selected epoch and temperature.

\begin{table}[!t]
\centering\small
\setlength{\tabcolsep}{1.45pt}
\renewcommand{\arraystretch}{1.10}
\begin{tabular}{@{}lrrrrrr@{}}
\toprule
\rowcolor{TableHeader} & & \multicolumn{3}{c}{\textbf{Test results}} & \multicolumn{2}{c@{}}{\textbf{Validation selection}} \\
\cmidrule(lr){3-5}\cmidrule(l){6-7}
Target & Seed & NLL & Acc. (\%) & Wrong (\%) & Epoch & $T$ \\
\midrule
\multicolumn{7}{@{}l@{}}{\textbf{Qwen3.5-4B}} \\
Unadapted & -- & 1.168 & 71.2 & 16.6 & -- & 1.41 \\
Human & 0 & 0.904 & 81.2 & 21.2 & 3 & 1.06 \\
Human & 1 & 0.907 & 81.2 & 22.7 & 2 & 1.06 \\
Human & 2 & 0.910 & 81.0 & 18.9 & 2 & 0.94 \\
Correct & 0 & 0.988 & 81.0 & 20.4 & 1 & 1.78 \\
Correct & 1 & 0.979 & 79.8 & 25.7 & 2 & 2.00 \\
Correct & 2 & 0.972 & 80.2 & 18.5 & 2 & 1.68 \\
\midrule
\multicolumn{7}{@{}l@{}}{\textbf{Intern-Decision-4B}} \\
Unadapted & -- & 1.132 & 72.3 & 15.6 & -- & 1.78 \\
Human & 0 & 0.905 & 81.5 & 24.4 & 3 & 1.06 \\
Human & 1 & 0.909 & 81.0 & 22.3 & 2 & 1.06 \\
Human & 2 & 0.911 & 81.0 & 19.9 & 2 & 0.94 \\
Correct & 0 & 0.995 & 79.0 & 16.6 & 1 & 1.78 \\
Correct & 1 & 0.984 & 81.2 & 18.5 & 2 & 2.00 \\
Correct & 2 & 0.971 & 80.8 & 18.6 & 2 & 1.78 \\
\midrule
\multicolumn{7}{@{}l@{}}{\textbf{LFM2.5-VL-3B}} \\
Unadapted & -- & 1.073 & 77.1 & 17.3 & -- & 2.52 \\
Human & 0 & 0.888 & 82.5 & 23.0 & 2 & 1.06 \\
Human & 1 & 0.889 & 83.8 & 22.5 & 2 & 1.06 \\
Human & 2 & 0.893 & 83.5 & 22.3 & 2 & 1.00 \\
Correct & 0 & 0.923 & 81.0 & 21.8 & 1 & 1.41 \\
Correct & 1 & 0.915 & 82.1 & 19.0 & 1 & 1.41 \\
Correct & 2 & 0.917 & 83.3 & 20.0 & 1 & 1.50 \\
\midrule
\multicolumn{7}{@{}l@{}}{\textbf{d1-3B}} \\
Unadapted & -- & 1.025 & 74.6 & 19.4 & -- & 1.12 \\
Human & 0 & 0.891 & 81.5 & 24.2 & 2 & 1.06 \\
Human & 1 & 0.889 & 82.3 & 25.6 & 2 & 1.06 \\
Human & 2 & 0.899 & 82.1 & 23.9 & 2 & 1.00 \\
Correct & 0 & 0.943 & 81.5 & 21.6 & 1 & 1.41 \\
Correct & 1 & 0.929 & 81.5 & 24.4 & 1 & 1.50 \\
Correct & 2 & 0.927 & 83.3 & 23.7 & 1 & 1.41 \\
\bottomrule
\end{tabular}
\caption{Within-ImageNet16H test point estimates for all 24 trained models and four unadapted checkpoints. Lower NLL is better, and higher accuracy and wrong-category agreement are better. Epochs and temperatures use only the within-dataset validation split. Unadapted checkpoints are shared across the three training seeds. Wrong denotes wrong-category agreement.}
\label{tab:within16h-arms}
\end{table}

\paragraph{Metrics and uncertainty.}\leavevmode\newline
We report trial-weighted human-choice NLL, image-equal accuracy and wrong-category agreement with the original definitions. Paired 95\% percentile intervals use 2,000 shared resamples of original sources with all phase variants together and of observers. The intervals condition on the fitted models and temperatures, and seeds are reported separately. Per-observer--condition groups have a median of five test trials. Source resampling makes these small groups more often exhibit perfect correctness agreement, shifting the per-group EC distribution away from its point estimate. We therefore omit EC for this control and retain its original numerical records separately. This does not alter the main experiment's metrics.

\paragraph{Human versus correct-label supervision.}\leavevmode\newline
Human supervision has lower NLL in 10 of 12 paired intervals, with the other two intervals containing zero (Table~\ref{tab:within16h-supervision}). All six Qwen/Intern comparisons favor human supervision, as do four of six LFM2/d1 comparisons. The two zero-containing cases are LFM2 seeds 1 and 2. All 12 accuracy contrasts and all 12 wrong-category-agreement contrasts contain zero. Thus the within-experiment NLL advantage does not imply an established advantage in accuracy or wrong-category agreement.

\begin{table}[!t]
\centering\small
\setlength{\tabcolsep}{3pt}
\renewcommand{\arraystretch}{1.04}
\begin{tabular}{@{}rrrr@{}}
\toprule
\rowcolor{TableExternal!70}Seed & $\Delta$NLL & $\Delta$Accuracy (pp) & $\Delta$Wrong (pp) \\
\midrule
\multicolumn{4}{@{}l@{}}{\textbf{Qwen3.5-4B}} \\
0 & $-0.084$ & $+0.2$ & $+0.7$ \\
 & {\footnotesize $[-0.122,-0.045]$} & {\footnotesize $[-2.9,+3.1]$} & {\footnotesize $[-9.5,+11.7]$} \\[1pt]
1 & $-0.072$ & $+1.5$ & $-3.1$ \\
 & {\footnotesize $[-0.108,-0.037]$} & {\footnotesize $[-1.5,+4.6]$} & {\footnotesize $[-11.5,+4.2]$} \\[1pt]
2 & $-0.062$ & $+0.8$ & $+0.4$ \\
 & {\footnotesize $[-0.097,-0.028]$} & {\footnotesize $[-2.1,+4.0]$} & {\footnotesize $[-9.6,+8.8]$} \\[1pt]
\midrule
\multicolumn{4}{@{}l@{}}{\textbf{Intern-Decision-4B}} \\
0 & $-0.090$ & $+2.5$ & $+7.8$ \\
 & {\footnotesize $[-0.133,-0.048]$} & {\footnotesize $[-0.8,+6.0]$} & {\footnotesize $[-0.6,+16.5]$} \\[1pt]
1 & $-0.075$ & $-0.2$ & $+3.8$ \\
 & {\footnotesize $[-0.112,-0.038]$} & {\footnotesize $[-2.9,+2.5]$} & {\footnotesize $[-3.6,+11.6]$} \\[1pt]
2 & $-0.060$ & $+0.2$ & $+1.3$ \\
 & {\footnotesize $[-0.096,-0.024]$} & {\footnotesize $[-3.1,+3.3]$} & {\footnotesize $[-9.0,+10.2]$} \\[1pt]
\midrule
\multicolumn{4}{@{}l@{}}{\textbf{LFM2.5-VL-3B}} \\
0 & $-0.035$ & $+1.5$ & $+1.2$ \\
 & {\footnotesize $[-0.062,-0.007]$} & {\footnotesize $[-0.6,+3.5]$} & {\footnotesize $[-6.5,+8.3]$} \\[1pt]
1 & $-0.026$ & $+1.7$ & $+3.5$ \\
 & {\footnotesize $[-0.049,+0.0001]$} & {\footnotesize $[-0.8,+4.2]$} & {\footnotesize $[-5.1,+12.1]$} \\[1pt]
2 & $-0.024$ & $+0.2$ & $+2.3$ \\
 & {\footnotesize $[-0.048,+0.001]$} & {\footnotesize $[-2.5,+2.9]$} & {\footnotesize $[-8.4,+13.9]$} \\[1pt]
\midrule
\multicolumn{4}{@{}l@{}}{\textbf{d1-3B}} \\
0 & $-0.052$ & $+0.0$ & $+2.6$ \\
 & {\footnotesize $[-0.079,-0.023]$} & {\footnotesize $[-2.3,+2.1]$} & {\footnotesize $[-5.5,+11.7]$} \\[1pt]
1 & $-0.040$ & $+0.8$ & $+1.2$ \\
 & {\footnotesize $[-0.066,-0.013]$} & {\footnotesize $[-1.5,+2.9]$} & {\footnotesize $[-6.1,+8.5]$} \\[1pt]
2 & $-0.028$ & $-1.3$ & $+0.2$ \\
 & {\footnotesize $[-0.053,-0.002]$} & {\footnotesize $[-3.5,+1.0]$} & {\footnotesize $[-7.9,+8.0]$} \\[1pt]
\bottomrule
\end{tabular}
\caption{Human-minus-correct-label supervision contrasts for every model and seed. Entries give the point difference and paired 95\% interval from the same 2,000 source/observer resamples. Negative NLL favors human supervision. Positive accuracy and wrong-category agreement favor human supervision. Secondary metrics use percentage points (pp).}
\label{tab:within16h-supervision}
\end{table}

\newpage
\paragraph{Checkpoint gaps before and after adaptation.}\leavevmode\newline
Before adaptation, the validation-calibrated Intern-minus-Qwen NLL gap is $-0.0363$ $[-0.0670,-0.0043]$, and the d1-minus-LFM2 gap is $-0.0485$ $[-0.0857,-0.0109]$. After human-choice adaptation all six decision-minus-base NLL intervals contain zero. After correct-label adaptation five contain zero, while d1 is worse than LFM2 at seed 0 with a gap of $+0.0201$ $[+0.0040,+0.0371]$.

\newpage
\raggedbottom
\noindent The directly paired change $J$ is positive in 11 of 12 intervals, and the Intern/Qwen correct-label seed-2 interval contains zero (Table~\ref{tab:within16h-checkpoint}). Positive $J$ denotes a smaller or reversed relative advantage, not necessarily a worsening of the decision checkpoint itself. Zero-containing final gaps do not establish equivalence. The corresponding secondary-metric changes are given in Table~\ref{tab:within16h-secondary}.

\begin{table}[t]
\centering\footnotesize
\setlength{\tabcolsep}{2.6pt}
\renewcommand{\arraystretch}{1.20}
\begin{tabular}{@{}lrrr@{}}
\toprule
\rowcolor{TableExternal!70}Target & Seed & After ($D_{\mathrm{adapt}}$) & Change ($J$) \\
\midrule
\multicolumn{4}{@{}l@{}}{\textbf{Intern/Qwen}\enspace Before ($D_0$)~$-0.036\;[-0.067,-0.004]$} \\
Human & 0 & $+0.001\;[-0.010,+0.012]$ & $+0.037\;[+0.004,+0.071]$ \\
Human & 1 & $+0.003\;[-0.008,+0.012]$ & $+0.039\;[+0.005,+0.072]$ \\
Human & 2 & $+0.001\;[-0.008,+0.010]$ & $+0.037\;[+0.004,+0.069]$ \\
Correct & 0 & $+0.007\;[-0.008,+0.023]$ & $+0.044\;[+0.011,+0.075]$ \\
Correct & 1 & $+0.005\;[-0.018,+0.029]$ & $+0.042\;[+0.002,+0.079]$ \\
Correct & 2 & $-0.001\;[-0.015,+0.011]$ & $+0.035\;[-0.001,+0.070]$ \\
\midrule
\multicolumn{4}{@{}l@{}}{\textbf{d1/LFM2}\enspace Before ($D_0$)~$-0.048\;[-0.086,-0.011]$} \\
Human & 0 & $+0.004\;[-0.011,+0.019]$ & $+0.052\;[+0.013,+0.089]$ \\
Human & 1 & $+0.000\;[-0.015,+0.016]$ & $+0.049\;[+0.011,+0.085]$ \\
Human & 2 & $+0.006\;[-0.008,+0.022]$ & $+0.055\;[+0.017,+0.093]$ \\
Correct & 0 & $+0.020\;[+0.004,+0.037]$ & $+0.069\;[+0.028,+0.109]$ \\
Correct & 1 & $+0.014\;[-0.003,+0.033]$ & $+0.063\;[+0.021,+0.104]$ \\
Correct & 2 & $+0.011\;[-0.009,+0.030]$ & $+0.059\;[+0.016,+0.102]$ \\
\bottomrule
\end{tabular}
\caption{Within-ImageNet16H decision-minus-base NLL gaps before adaptation ($D_0$), after adaptation ($D_{\mathrm{adapt}}$), and their directly paired change $J=D_{\mathrm{adapt}}-D_0$. Entries are differences with paired 95\% intervals. Negative gaps favor the decision checkpoint. Positive $J$ indicates a reduction or reversal of its initial advantage. Each pair shares one unadapted gap across targets and seeds.}
\label{tab:within16h-checkpoint}
\end{table}

\paragraph{Relation to the transfer result.}
Human supervision improves NLL relative to correct labels in most within-experiment comparisons, whereas its NLL benefit does not reliably transfer from model-vs-human to ImageNet16H. The training datasets and image splits differ between these protocols.

\newpage
\noindent Their contrast is consistent with a dependence on the training experiment, but it does not isolate the contributions of images, observers or procedure. The reduction in the decision checkpoint's starting advantage occurs under both supervision targets in this within-experiment control.

\medskip
\noindent\begin{minipage}{\columnwidth}
\centering\small
\setlength{\tabcolsep}{3pt}
\renewcommand{\arraystretch}{1.25}
\begin{tabular}{@{}lrrr@{}}
\toprule
\rowcolor{TableExternal!70}Target & Seed & Accuracy $J$ (pp) & Wrong choice $J$ (pp) \\
\multicolumn{4}{@{}l@{}}{\footnotesize $J>0$ increases the decision-model advantage.} \\
\midrule
\multicolumn{4}{@{}l@{}}{\textbf{Intern/Qwen}} \\
Human & 0 & $-0.8\;[-2.9,+1.3]$ & $+4.1\;[-3.0,+12.1]$ \\
Human & 1 & $-1.2\;[-3.7,+1.5]$ & $+0.6\;[-9.9,+8.8]$ \\
Human & 2 & $-1.0\;[-2.9,+0.8]$ & $+2.0\;[-4.7,+8.6]$ \\
Correct & 0 & $-3.1\;[-5.4,-1.0]$ & $-2.9\;[-12.3,+6.2]$ \\
Correct & 1 & $+0.4\;[-2.5,+3.7]$ & $-6.3\;[-19.2,+4.2]$ \\
Correct & 2 & $-0.4\;[-2.7,+2.1]$ & $+1.1\;[-6.0,+7.4]$ \\
\midrule
\multicolumn{4}{@{}l@{}}{\textbf{d1/LFM2}} \\
Human & 0 & $+1.5\;[-1.0,+4.0]$ & $-1.0\;[-9.9,+7.6]$ \\
Human & 1 & $+1.0\;[-1.5,+3.5]$ & $+0.9\;[-8.1,+9.3]$ \\
Human & 2 & $+1.0\;[-1.2,+3.3]$ & $-0.5\;[-9.4,+7.8]$ \\
Correct & 0 & $+2.9\;[+0.4,+5.4]$ & $-2.4\;[-11.6,+5.4]$ \\
Correct & 1 & $+1.9\;[-0.8,+4.6]$ & $+3.3\;[-6.6,+12.7]$ \\
Correct & 2 & $+2.5\;[-0.2,+5.0]$ & $+1.6\;[-8.3,+11.2]$ \\
\bottomrule
\end{tabular}
\captionof{table}{Secondary-metric changes in the decision-minus-base gap for all targets and seeds, with paired 95\% intervals. Here positive values indicate an increased decision-model advantage, the opposite direction to NLL. Percentage points are denoted pp. Zero-containing and opposite-direction results are retained.}
\label{tab:within16h-secondary}
\end{minipage}\par

\newpage

\raggedend
\section{Observer-history controls}
Three history-conditioned fits share 70,131 training, 7,454 validation and 5,078 test current-target trials. The own-history input contains the four preceding valid image--response trials of the same observer in the same split and session. The other-history input uses one donor observer's responses to the same four images, and the no-history input omits them. Targets without four eligible predecessors or a matched donor are excluded from all arms, and the current response never appears in the input. Source validation selects epoch 2 and temperature 1 for all three arms.

Own-history NLL is 1.4672, against 1.4773 without history and 1.4815 with another observer's history (Table~\ref{tab:history}). Own minus no history is $-0.0101$ $[-0.0195,-0.0009]$, and own minus other history is $-0.0143$ $[-0.0212,-0.0072]$. All secondary-metric intervals for both contrasts contain zero. An earlier inference-only diagnostic, which reused a group-trained Qwen adapter without history training, was adverse, with own minus no history at $+0.6435$ NLL for one observer. The protocol never repeats a target image for the same observer, so these results cannot separate stable preferences from momentary changes, trial order or the information carried by the previous images. A matched decision-checkpoint history comparison, further conditions and more training seeds are not yet available.

\medskip
\noindent\begin{minipage}{\columnwidth}
\centering\small
\begin{tabular}{lrrrr}
\toprule
\rowcolor{TableHeader}\textbf{Arm} & \textbf{NLL} $\downarrow$ & \textbf{EC} $\uparrow$ & \textbf{Wrong} $\uparrow$ & \textbf{Acc.} $\uparrow$\\\midrule
no history & 1.4773 & 0.3239 & 0.2030 & 0.5977\\
own history & \textbf{1.4672} & \textbf{0.3448} & \textbf{0.2211} & 0.6032\\
other history & 1.4815 & 0.3357 & 0.2110 & \textbf{0.6060}\\
\bottomrule\end{tabular}
\captionof{table}{Trained-history point estimates. Wrong denotes wrong-category agreement. Bold marks the best point estimate in each column. All paired 95\% intervals are retained in the numerical archive.}
\label{tab:history}
\end{minipage}\par

\section{Extended related work}
\label{sec:relatedextended}
Human visual alignment involves several distinct comparisons. Controlled image degradations and conflicts between texture and shape reveal differences between human and CNN behavior~\cite{geirhos2018generalisation,geirhos2019texture}. Model-vs-human provides 17 visual datasets and laboratory human comparisons~\cite{geirhos2021partial}, while EC measures shared trial-level successes and failures~\cite{geirhos2020beyond}. HuDec also scores probabilities assigned to particular human labels and agreement on the category when both systems are wrong.

Visual diagnostic suites target complementary capabilities. Cai et al. test contrast responses using image-encoder feature distances~\cite{cai2025lowlevel}. DepthCues probes monocular depth cues in frozen representations~\cite{danier2025depthcues}, and contour-fragmentation experiments examine object recognition under controlled visual evidence~\cite{lonnqvist2025contour}. HVSBench evaluates MLLM behavior across prominence, subitizing, prioritizing, free viewing and search~\cite{lin2024hvsbench}.

\newpage
\noindent Probe-readout performance and standardized textual perceptual responses do not directly measure HuDec's fixed-category human-choice likelihood.

Human-label training predates modern vision-language models. Peterson et al. study human uncertainty labels and class-average distributions~\cite{peterson2019human}. Schulze Buschoff et al. examine visual-cognition fine-tuning and transfer~\cite{schulzebuschoff2025testing}. AligNet transfers human-aligned similarity structure from a teacher into visual representations, improving similarity judgments and downstream generalization~\cite{muttenthaler2025aligning}. HuDec studies degraded-image category choices and asks how the advantage of a released decision checkpoint changes when both members of its pair learn the same human responses.

Probability and attentional alignment also require separate treatment. Lan et al. distinguish accuracy-oriented calibration from calibration toward human response distributions in VQA~\cite{lan2025uncertainty}. Hong et al. show that cross-modal human-choice supervision can improve choice agreement without matching human attention~\cite{hong2026similar}. Their two-choice association task differs from HuDec's object recognition, but motivates separating predictive behavioral agreement from mechanistic interpretation.

Centaur predicts choices from task descriptions and participant histories~\cite{binz2025foundation}. Psych-201 reports that assistant post-training can reduce alignment on language-based behavioral tasks~\cite{binz2026post}. Chang Ortega et al. vary generative and discriminative objectives and find task-dependent human alignment, including model-vs-human evaluation~\cite{changortega2026sweetspot}. This differs from tracking released base and decision pairs before and after matched human-choice adaptation. Our history analysis conditions on recent observations of images and responses and does not infer stable individual traits.

Human-referenced cognitive benchmarks provide complementary evidence. CogBattery shows that matching human condition effects need not match errors or confidence reports~\cite{hou2026cogbattery}. S1CogBench evaluates System One models on established cognitive tasks~\cite{hou2026s1cogbench}. EEGJev uses a System One model to score candidate answers to questions about EEG recordings~\cite{hou2026eegjev}. HuDec combines actual visual inputs, checkpoint pairing, human supervision and cross-experiment transfer. Human soft targets, human fine-tuning and transfer failures are not individually new contributions.

Our OpenCLIP ViT-H/14 and low-pass baselines follow Wolff et al.~\cite{wolff2026lowpass} in model and blur, but they use our held-out splits and calibration, so their scores are not comparable to the published 17-dataset aggregate.

Broader adaptation work shows that full fine-tuning can yield lower shifted-distribution accuracy than linear probing despite higher in-distribution accuracy~\cite{kumar2022finetuning}, and that interpolating pretrained and adapted weights can preserve robustness~\cite{wortsman2022robust}. These are classification-transfer results, whereas HuDec measures human-choice likelihood and paired checkpoint changes under a fixed adaptation recipe.

\end{document}